# A Chinese Text Classification Method With Low Hardware Requirement Based on Improved Model Concatenation

Qingli Man[1], Yuanhao ZHUO[1,2]

[1]Zhiwei Research Institute, Ningbo, China

[2] School of Information Technology and Computer Science, University of Melbourne

manqingli@zhiweidata.com, yuanhaozhuo@gmail.com

**Abstract:** In order to improve the accuracy performance of Chinese text classification models with low hardware requirements, an improved concatenation-based model is designed in this paper, which is a concatenation of 5 different sub-models, including TextCNN, LSTM, and Bi-LSTM. Compared with the existing ensemble learning method, for a text classification mission, this model's accuracy is 2% higher. Meanwhile, the hardware requirements of this model are much lower than the BERT-based model.

## 1. Introduction

The invention of the transformer and transformer-based BERT resulted in the rapid development of natural language processing (NLP) in recent years. BERT-related models have demonstrated strong capabilities in various tasks of NLP like text classification, translation, Q&A[1]. As the most spoken language, Chinese NLP has its important value. Designed for English text, BERT can also have excellent performance on Chinese text after adjustment, and this is an extremely important step in the development of Chinese NLP.

However, the transformers have extremely high requirements for hardware performance. In actual industrial production, data companies often need to face millions or even more data. Such a large amount of data itself will bring a great burden to the hardware, and the training of deep learning models makes the hardware cost of the enterprise even higher. As a result, small businesses or groups usually cannot afford the high hardware costs, thus they cannot train or apply the latest models.

Aiming at the situation that start-up data companies or small groups cannot afford high-performance hardware, this paper designs a model based on an improved model concatenation method that can run via 16G memory and NVIDIA GTX1070Ti graphics card, and this model has good results in practical Chinese text classification applications and has a better performance compared with the most commonly used ensemble learning voting method.

## 2. Related works

To achieve the best training performance with the lowest hardware requirements, first, consider the most commonly used RNN model before the appearance of the transformer. Here, consider the Bi-directional Long Short-Term Memory (Bi-LSTM) model designed by Wang and his college[2] as a comparison group.

Another deep learning model known for its fast speed and low hardware requirements is the Text Convolutional Neural Network (TextCNN) model. Unlike image processing in Computer vision (CV), CNN for NLP can be designed as 1-D or 2-D according to demand. Here, consider the TextCNN model designed by Wang and his college[3] as a comparison group.

In CV, deep CNN can speed up training through multiple pooling. This type of network is known as VGG. The essential reason for the acceleration of VGG is the reduced number of parameters caused by pooling, so this can also avoid the problem of excessive hardware requirements for deep networks. Here, consider the VGG10 model as a comparison group.

In order to fully use the advantages of a variety of different models, and maintain the low requirements for hardware performance, a common method is ensemble learning[4], which is based on different submodels. Consider the most commonly used weighted voting method as a comparison group.

Besides, the ALBERT Chinese model released by Google in 2019 is also selected, which is a simplified version of BERT[1,5] as a comparison group to show that our concatenation-based model is better than BERT in hardware requirements.

All these comparison results will be shown in the Result and Conclusion chapter.

# 3. Methodology

The existing ensemble learning is based on multiple sub-models that have been trained, and further operations such as weighted voting are performed in the follow-up. This method is either extremely easy to over-fit during the second training, or, like weighted voting, the acquisition of weights is completely independent of the training process. Therefore, instead of voting outside the model, our concatenation-based model uses concatenate inside one model to ensemble different sub-models. The concatenation-based model structure is shown in Figure 1.

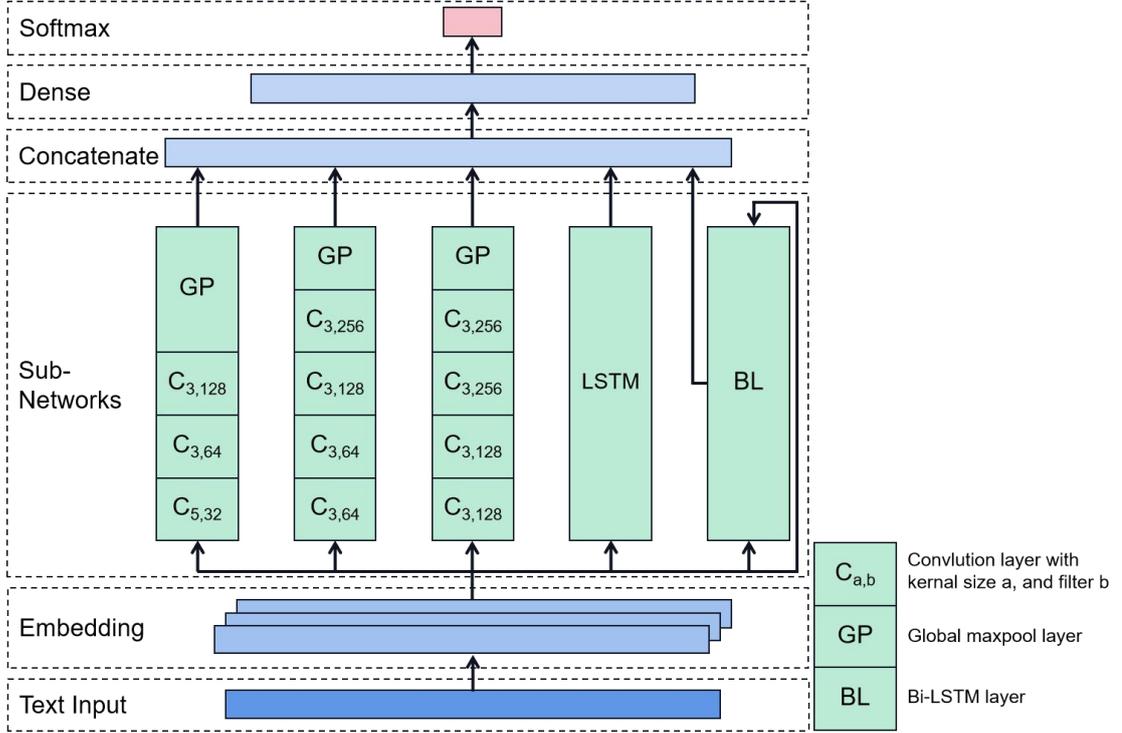

Figure 1: Overall concatenation-based model structure

For the input text, the first part of our concatenation-based model is the embedding. Considering that the purpose of our model is to reduce the hardware performance requirements as much as possible, although it is Chinese text classification, in our model, there is still no word segmentation, and Chinese character-level input is selected, and then the look-up table is used to convert each character into a numeric value, and then perform random initial value embedding. Since the Chinese model of word segmentation often requires a vocabulary of hundreds of thousands of lengths, there are only a few thousand Chinese characters without word segmentation, so this can greatly reduce the vocabulary size. At the same time, unlike English letters, Chinese characters are hieroglyphs, and each Chinese character itself contains a lot of information. As a result, the unsegmented model will not influence the final accuracy.

As shown in Figure 1, after embedding, the data was fed into 5 sub-networks, including 3 TextCNN networks with different parameters, an LSTM network, and a Bi-LSTM network.

For the TextCNN sub-network, each sub-network contains several CNN layers, as well as a global pooling layer. The role of the global pooling layer is to replace the pooling layer with a global pooling layer with the same size as the entire feature map, and it is also capable of resizing the tensor to replace the flatten layer. In my model, the pooling layer is not used in the TextCNN sub-network. This is because the pooling layer does not work well in NLP. This can also be reflected in the poor performance of VGG in the results shown in the conclusion chapter. Since the parameters of each TextCNN sub-network are different, the features obtained by each TextCNN are different, namely each sub-network can provide different features for the subsequent. The parameters of each TextCNN sub-network is shown in figure 1, and the activation function used for every CNN layer is relu. For input $x$, the output of the convolution layer can be written as[3]:

$$C = relu(K \times x_{i:i+h-1} + b) \quad (1)$$

Where $relu(\ )$ is the function for relu, $K$ is the convolution kernel, and $b$ is an offset value. The input $x$ can be the result of the embedding layer, or the output of the previous convolution layer. The results of the last convolution layer are finally output after passing a global pooling block.

For the LSTM sub-network and Bi-LSTM sub-network, the simple LSTM is an optimized version of RNN to solve the problem of gradient disappearance[2]. Our concatenation-based model selects LSTM and Bi-LSTM with 250 hidden neurons, and the activation function is tanh. Take Bi-LSTM as an example, the figure 2 is its structure, which is a combination of a leftward LSTM and a rightward LSTM. The output of the Bi-LSTM layer can be written as:

$$leftward : \vec{h}_t = H(\vec{W}x_t + \vec{V}\vec{h}_{t-1} + b_{\vec{h}}) \quad (2)$$

$$rightward : \vec{h}_t = H(\vec{W}x_t + \vec{V}\vec{h}_{t-1} + b_{\vec{h}}) \quad (3)$$

$$y_t = \vec{W}_y \times \vec{h}_t + \vec{W}_y \times \vec{h}_t + b_y \quad (4)$$

Where $W$ and $V$ are coefficients of a linear relationship, $H$ is the LSTM block transition function.

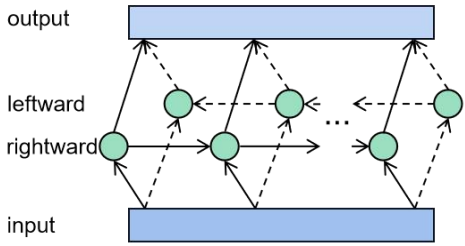

Figure 2: The structure of the Bi-LSTM model

After obtaining the results of each sub-network, the next step is to concatenate the output of each sub-network. The output of each sub-network contains the features of the data learned by this sub-network. Some of these features are valid and strong. They can help the entire model find the correct results, while a large part of them are invalid or weak features. The purpose of this step is to integrate these features and concatenate the features obtained from each network in series to obtain as many effective features as possible. After passing through the subsequent fully connected network, our method theoretically can help to improve the accuracy of the entire model, and compared to the voting ensemble learning model where the weights are artificially given and cannot be trained, the parameters of this model have similar effects while their values are obtained in a fully connected network, and can be trained, so theoretically it can have better effects than the ensemble model.

The last part of the concatenation-based model is the fully connected layer and softmax output. This part is relatively conventional, so I won't go into detail.

## 4. Experiments and Results

### 4.1 Data Set

The data set used in this article is a data set used in actual industrial production. The text source is a variety of Chinese channels obtained through crawlers, including but not limited to Weibo, blogs, self-media, portals, and mobile applications Tweets. The original data set contains approximately 6 million pieces of data belonging to 25 categories (including society, finance, international, sports, technology, current events, etc.), the shortest text contains one Chinese character, and the longest text is approximately 500,000 Chinese characters.

To reduce the hardware burden, we gave up text with a text length of more than 5000 Chinese characters. For text with a text length smaller than 10, these texts are mostly emoji or abbreviations in specific contexts, therefore text with a length smaller than 10 will be abandoned. For text with a text length of fewer than 5000 characters and more than 10, it is divided into long text and short text with 500 characters as the limit. Based on this grouping, three data sets in the following table were used in subsequent experiments:

Table 1: Data sets structure

| Data set# | Max length | Min length | categories | Train set size |
|---|---|---|---|---|
| 1 | 5000 | 10 | 25 | 25×30000 |
| 2 | 500 | 10 | 25 | 25×10000 |
| 3 | 5000 | 500 | 25 | 25×20000 |

The first data set is a complete data set, including long text and short text, to simulate the universal situation and get a universal model.

The second data set is a short text data set, which only contains short texts whose length is less than 500. These data are mainly informal and mainly come from Weibo and Bullet screen.

The third data set is a long text data set, containing only text longer than 500, and is used to train a model aim at long texts. These data are mainly formal articles and mainly come from blogs, self-media, and portals.

### 4.2 Baseline Methods

To reflect the effectiveness of our concatenation-based model, based on related work, we designed several other groups of models as a control group to compare results. They are:
- **TextCNN**[3]: The most basic model for deep learning. The

model used in this experiment is 2 layer TextCNN with a kernel size of 3, and 128 filters for each layer.
- **Bi-LSTM**[2]: The most commonly used RNN model for NLP. The model used in this experiment is Bi-LSTM with hidden neurons of 250.
- **VGG**: A CNN network with fewer parameters and faster calculation speed. VGG10 is used in this experiment.
- **ALBERT+TextCNN**[1,5]: A model with a much higher accuracy rate, but extremely high hardware requirements. The TextCNN part is a 3-layer regular TextCNN.
- **Ensemble Learning**[4,6]: An existing method of improving accuracy without high hardware requirements. Use the weight voting method to ensemble three basic models (2 TextCNN with different parameters, and a Bi-LSTM).

## 4.3 Experimental Result

For each group of models, we adopted the accuracy for the test set as the evaluation index.

For the first data set, the results of each group of models are shown in Table 2:

Table 2: Experimental Result for Data set 1

(using NVIDIA GTX1070Ti)

|  | Model | Test_acc |
|---|---|---|
| Data set 1 (long&short) | Text CNN | 74.04% |
|  | Bi-LSTM | 74.47% |
|  | VGG | 70.30% |
|  | ALBERT+CNN | OOM |
|  | Ensemble Learning | 78.60% |
|  | Our model | **80.99%** |

For the second data set, the results of each group of models are shown in Table 3:

Table 3: Experimental Result for Data set 2

(using NVIDIA GTX1070Ti)

|  | Model | Test_acc |
|---|---|---|
| Data set 2 (short) | Text CNN | 73.15% |
|  | Bi-LSTM | 74.62% |
|  | VGG | 69.81% |
|  | ALBERT+CNN | OOM |
|  | Ensemble Learning | 78.05% |
|  | Our model | **79.13%** |

For the third data set, the results of each group of models are shown in Table 4:

Table 4: Experimental Result for Data set 3

(using NVIDIA GTX1070Ti)

|  | Model | Test_acc |
|---|---|---|
| Data set 3 (long) | Text CNN | 73.58% |
|  | Bi-LSTM | 73.67% |
|  | VGG | 70.14% |
|  | Ensemble Learning | 77.96% |
|  | Our model | **80.12%** |

Since ALBERT+TextCNN has already shown out of memory (OOM) results in the data set with a smaller memory requirement, there is no more analysis in the last part.

From the results of the three data sets, it is not difficult to find that our concatenation-based model and Ensemble Learning have significantly improved the test accuracy compared to simple CNN and LSTM, while the CNN and LSTM perform better than VGG.

Specifically compare the three improved models of Bert, Ensemble Learning (EL), and our model. The following table shows the operability of each model, the average test accuracy, and the time needed for 30000 data prediction (Include the time to read the model but not include the time to read the data).

Table 5: Comparison of three optimized models

(using NVIDIA GTX1070Ti)

| Model | Operability via GTX 1070Ti | Average Test accuracy | Time for 30000 data prediction |
|---|---|---|---|
| BERT-based | Inoperable | Meaningless | Meaningless |
| EL | Operable | 78.20% | 89 seconds |
| Our model | Operable | 80.08% | 175 seconds |

Although all three models can run via Google's Colab, in the case of insufficient hardware performance, the BERT-based model will be OOM on a computer with RTX1070Ti graphics card and 12G memory, and no effective results can be obtained. Since our goal is to find an algorithm with lower hardware performance requirements, the BERT-based model does not meet our requirements well at this point, so we did not do further accuracy analysis.

Compared with Ensemble Learning, it is obvious from the results shown in the table that our concatenation-based model has an approximately 2% improvement in the accuracy of the test set. And our model successfully met the conditions for running on relatively low-performance hardware. The drawback of our model is this model needs more time for prediction compared with Ensemble Learning.

# 5. Conclusions

From the results obtained above, it is not difficult to find that our concatenation-based model can run under lower hardware requirements compared to the Bert-based model. Compared with Ensemble Learning, our model has a certain improvement in accuracy (about 2% improvement in the data set we tested). This is because compared to the Bert-based model, the number of parameters in the concatenation-based model is smaller, which makes the smaller requirement of memory. Compared with the Ensemble Learning voting method where the weight is set artificially, and not trainable, the concatenation-based model uses a trainable fully connected network to get the results, making the results more accurate.

A weakness of the concatenation-based model is that compared to Ensemble Learning, our model has a slower prediction speed, which is also a place to be improved in the future.

Overall, this article proposes a model that can have a higher accuracy rate under low hardware requirements compared with existing models.